\documentclass[10pt, conference, compsocconf]{IEEEtran}
\ifCLASSINFOpdf
\else
\fi
\usepackage{algpseudocode}
\usepackage{algorithm}
\usepackage{multirow}
\usepackage{amsfonts}
\usepackage{bbold}

\usepackage{amsthm}
\usepackage{subfig}
\usepackage[pdftex]{graphicx}
\usepackage[cmex10]{amsmath}
\usepackage[final]{pdfpages} 
\usepackage{url}
\usepackage{microtype}
\usepackage{caption}

\newcommand{\argmax}{\operatornamewithlimits{argmax}}

\newtheorem*{MODL}{Definition}
\newtheorem*{informativity}{Definition}
\newtheorem*{MI1}{Definition}
\newtheorem*{MI2}{Definition}
\newtheorem*{diffcrit}{Theorem}

\begin{document}
%
\title{A Triclustering Approach for Time Evolving Graphs}



%

\author{
\IEEEauthorblockN{Romain Guigour\`es \IEEEauthorrefmark{1}\IEEEauthorrefmark{2}, Marc Boull\'e\IEEEauthorrefmark{1}, Fabrice Rossi\IEEEauthorrefmark{2}}

\IEEEauthorblockA{\IEEEauthorrefmark{1}Orange Labs, Lannion, France \\
\IEEEauthorrefmark{2}SAMM EA 4543, Universit\'e Paris 1, Paris, France \\
Email: romain.guigoures@orange.com, marc.boulle@orange.com, fabrice.rossi@univ-paris1.fr}
}


\maketitle

\begin{abstract}
This paper introduces a novel technique to track structures in time evolving
  graphs. The method is based on a parameter free approach for
  three-dimensional co-clustering of the source vertices, the target vertices
  and the time. All these features are simultaneously segmented in order to
  build time segments and clusters of vertices whose edge distributions are
  similar and evolve in the same way over the time segments. The main novelty
  of this approach lies in that the time segments are directly inferred from
  the evolution of the edge distribution between the vertices, thus not
  requiring the user to make an a priori discretization. Experiments conducted
  on a synthetic dataset illustrate the good behaviour of the technique, and a
  study of a real-life dataset shows the potential of the proposed approach
  for exploratory data analysis.
\end{abstract}


%
\IEEEpeerreviewmaketitle

%
\section{Introduction}
In real world problems, interaction between entities are generally
evolving through time. This is the case for instance in collaboration
networks between scientists when new PhD students are recruited or
conclude their thesis, when researchers move from one team to another,
etc. Understanding the corresponding time evolving interaction graphs
implies both to discover structures in those graphs and to track the
evolution of those structures through time. In this paper, we address
this problem by introducing a form of \emph{temporal blockmodeling}.

The concept of \emph{blockmodeling} originates in the pioneering works on
quantitative graph structure analysis conducted by sociologists in the
1950s in the context of social network analysis. Vertices
of the graph represent here social actors (also called subjects), while edges
between them correspond to social interaction. Among other topics,
sociologists were interested in structuring the interrelations between actors
into (social) roles \cite{Bott1957,Nadel1957}. This led in particular to the
introduction of the structural equivalence notion \cite{White1971}: two actors
are said to be \emph{structurally equivalent} if they play the same role in
the social network, that is if they interact in the same way with the same
actors. By grouping structurally equivalent actors, that is vertices in the
corresponding graph, one obtains a simplified and synthetic version of the
original graph. A generalization of structural equivalence was introduced
later on to relax its very strong constraints, under the name of
\emph{regular equivalence} \cite{White1983}. This consists in grouping actors
into clusters which interact identically with the same clusters.

To track the underlying structure of a graph, a matrix representation of a
graph is usually exploited, generally its adjacency matrix. Rows and columns
represent the actors, and the values of the matrix indicate whether there is a
relation between the actors represented. 
Early sociological approaches suggested to rearrange the rows and the column in order to partition the
matrix in homogeneous blocks, a technique called \emph{blockmodeling}.
Once the blocks are extracted, a partition of the subjects of both rows and
column can be made. This type simultaneous grouping is named
\emph{co-clustering}. A convenient way to represent the co-clustering is
through the \emph{image graph}: its vertices are the clusters of
subjects/actors identified in the blockmodeling; there is an edge between
two cluster vertices if there are edges between the actors that belong to
those clusters in the original social network (those edges are the ones that
define the characteristics of the roles in a case of regular equivalence).

Numerous methods have been proposed to build a satisfactory image graph. 
Some of them \cite{Doreian2004} are based on the
optimization of criteria that favor partitions with homogeneous blocks,
especially with pure zero-blocks as recommended in \cite{White1976}. More
recent deterministic approaches have focused on optimizing
criteria that quantify how well the image graph summarizes the graph
\cite{Reichardt2007} (see e.g. \cite{Wasserman1994} for details on such
criteria). Other approaches include \emph{stochastic} blockmodeling. In those
generative models, a latent cluster indicator variable is associated to each
actor/vertex. Conditionally to their latent variables, the probability of
observing an edge between two actors follows some standard distribution (a
Bernoulli distribution in the simplest case) whose parameters depend only on
the pair of clusters designated by the latent variables. In early approaches,
the number of clusters is chosen by the user \cite{Snijders2001}. More recent
techniques determine automatically the number of clusters using a
Dirichlet Process \cite{Kemp2006}. Finally, some recent approaches consider
non-boolean latent variables: cluster assignments are not strong and a vertex
has an affiliation degree to each cluster \cite{Airoldi2008}.%

Studies on evolving graphs are quite recent. The majority of the methods
define an evolving graph as a sequence of static snapshots. In some
approaches, the times segments are obtained by making an agglomerative
hierarchical grouping of the snapshots and intervals using a similarity
measure \cite{Hopcroft2004}. 
As for stochastic blockmodeling, an adaptation of the mixed membership is proposed that studies 
the evolution of the latent variables over time \cite{Xing2010}. 
Graphscope \cite{Faloutsos2007} is a two-stage method dedicated to simple bipartite graphs that tracks structures within time-evolving graphs. First, a partition of the snapshots is retrieved and evaluated using a MDL framework \cite{Grunwald2007}, then an agglomerative process is used to determine the temporal segmentation. 
As discussed in \cite{Lang2009}, the partitioning results may be sensitive to the coding schemes: in particular, coding shemes like those used in \cite{Faloutsos2007} have no guarantee of robustness w.r.t. random graphs.
The method we introduced in this paper is related to Graphscope in that it also exploits a modelization in blocks and is parameter free.
However, our approach exploits a robust modelization technique with high resiliance to noise (see Section~\ref{sec:expe}), can be applied to a large family of graphs (directed, undirected, bipartite, simple or multigraph) and considers simultaneously the partitioning of the graph and the discretization of the temporal variable within a global triclustering process, avoiding the use of a two-stage method.\\
In this paper, we propose a form of temporal blockmodeling for evolving graph
built upon the MODL approach \cite{Boulle2010}. As in classical blockmodeling,
our parameter free method groups vertices whose edges are similarly
distributed over the clusters. In addition, it partitions the time interval
into time segments during which the edge distributions between the clusters
are stationary. In order to obtain such a synthetic representation of the
evolving graph, a tri-clustering method is introduced. It optimizes
simultaneously the vertices co-clusters and the time interval partition. This
approach is resilient to noise and reliable in the senses that no
co-clustering structure is detected in case of random graphs and that no time
segmentation is made in case of globally stationary graphs. In addition the
true underlying distribution is asymptotically estimated. Section 2, describes
the proposed approach in details and introduces a post-treatment technique
useful as an exploratory analysis tool. Section 3 investigates the behaviour
of the method using an artificial dataset. Finally, the method is applied on a
real-life dataset in order to prove its effectiveness on a practical case.

\section{Evolving Graph Model}
We study graphs with multiple edges and therefore the method is not
restricted to binary or symetrical adjacency matrices. Let us denote an evolving graph $\mathcal{G} = \langle V_S,V_T,E(t)\rangle$ where the sets of vertices $V_S$ and $V_T$ are constant and $E(t)$ is the set of edges observed at time $t \in [T_{min}, T_{max}]$. This setting is general enough to account for simple graphs, multigraphs, directed graphs, bipartite graphs and undirected graphs, where each edge comes twice with the two directions.

\subsection{Model Definition}
As the graph edges are evolving through time, we replace the synthetic
representation by a unique image graph by a sequence of image graphs,
$\mathcal{IG}=(\mathcal{IG} _n)_{n=1,\ldots, N}$. Each image graph is supposed
to be a synthetic representation of the graph on a specific time segment. The
description of a graph and its image are displayed on Table
\ref{tab:imageGraph}.  

\vspace{-0.3cm}
\begin{table}[ht]
	\centering
	\footnotesize
		\begin{tabular}{|l|l|} 
		\hline
		\centering \textbf{Graph  $\mathcal{G}$} & \textbf{Image Graph  $\mathcal{IG}$} \\
		\hline
		$V_S$ set of source vertices &$C_S$ set of $K_S$ clusters of source vertices \\
		$V_T$ set of target vertices &$C_T$ set of $K_T$ clusters of target vertices \\
		$T$ the temporal variable &$I=\{I_1,I_2,...,I_N\}$ set of time segments\\
		$E(t \in T)$ evolving edges & $E_{\mathcal{IG}}(I_n)$ edges between clusters at $I_{n}$\\
		\hline
		\end{tabular}
	\caption{Data Representation of the initial graph and the image graph}
	\label{tab:imageGraph}
\end{table}
%
%

Now that the different components of an image graph are introduced, their parametrization must be specified. A model characterizing an image graph is defined by:
\begin{enumerate}
	\item the number of source and target clusters ($K_S$ and $K_T$);
	\item the number of time segments ($N$);
	\item the partition of the source vertices (resp. target vertices) into the source (resp. target) clusters of vertices;
	\item the distribution of the temporal edges of the graph on the co-clusters of source vertices, target vertices and time (i.e the edges of the image graph). Given this specification, we can derive from the graph the frequency of the clusters and time segments. Since time is a continuous variable, we can deduce the time segments bounds from their frequency;
	\item for each source (resp. target) cluster of vertices, the distribution of the edges whose source (resp. target) belongs to the cluster on the vertices of the cluster.
\end{enumerate}

Notice that specifications defined in third and fifth points are not required for the temporal variable.  
We require that a good temporal discretization should be invariant w.r.t. any monotonous transformation of the input time interval and robust w.r.t. atypical values (outliers). Given this requirement, we choose to exploit the ranks of the input values in the data sample rather than the values themselves. Thus, it is not necessary to specify how the timestamps are distributed over the time segments since time segments follow a logical order. As for the distribution of edges on the ranks locally to each interval, it is also not specified since we consider in our model that there is one rank per timestamp and thus only one way to distribute the edges over a time segment.

\subsection{MODL, the Criterion}

Given the model definition, the method we use is similar to a co-clustering with three features: the source and the target vertices are grouped and the time is discretized. In order to infer the best 3D partition, a criterion is built following a MAP (Maximum A Posteriori) Approach: $\mathcal{IG}^*=\argmax_{\mathcal{IG}}P(\mathcal{IG}) P(\mathcal{G}|\mathcal {IG})$. We detail below the two contributions $P(\mathcal{IG})$, a prior on the image graphs, and $P(\mathcal{G}|\mathcal{IG})$, the likelihood of the graph conditionally to an image graph.

\paragraph{Prior} Directly learning the model (image graph) on the data (graph) would enable the model to consider noisy phenomena as significative patterns and thus increases the risk of overfitting. To overcome this issue, a prior on the model penalizes the likelihood.

The prior is built hierarchically and uniformly at each stage in order to be uninformative \cite{Jaynes2003}. By doing this, we make no assumption on the data distribution. The following enumeration is a description of the a priori terms that constitutes the prior on the model:\\

\emph{(i)} The number of source clusters $K_S$ (resp. target clusters $K_T$) is
  uniformly distributed between $1$ and $|V_S|$, the number of source vertices
  (resp. $|V_T|$, the number of target vertices). The case with one single
  cluster corresponds to the null model, where there is no significative
  pattern within the graph. The other extreme case corresponds to the finest
  model where each vertex has a significant enough role to be clustered alone:
  the model has as many clusters as vertices . Both clustering structures are
  consistent with regular equivalence \cite{Borgatti1988,White1983}. Following
  the same idea, $N$, the number of time segments, is uniformly distributed
  between $1$ and the number of edges $|E|$. The case with one time segment
  corresponds to a stationary graph over time. The one with as many time
  segments as edges is an extremely fine-grained discretization: as time is a
  continuous variable, this case is allowed in our approach.  

\footnotesize
\[
p(K_S) = \dfrac{1}{|V_S|}\mbox{ ; }p(K_T) = \dfrac{1}{|V_T|} \mbox{ ; } p(N) =
\dfrac{1}{|E|}
\]
\normalsize

\emph{(ii)}  For a given number of source clusters (resp. target clusters), every partition of the $|V_S|$ vertices (resp. $|V_T|$ vertices) is equiprobable.

\footnotesize
\[p(\{C_S\} | K_S) = \dfrac{1}{B(|V_S|,K_S)} \mbox{ ; } p(\{C_T\} | K_T) = \dfrac{1}{B(|V_T|,K_T)}\]
\normalsize

where $B(|V_S|,K_S)=\sum_{k=1}^{K_S} S(|V_S|,k)$ is a sum of Stirling numbers of second kind, i.e the number of way of partitioning $|V_S|$ elements into $k$ non-empty subsets. At this step, no a priori hypothesis has been made on the clustering schemes. This point has been raised in \cite{Kemp2006} where a Dirichlet process is used as a prior on the number of clusters and on the distribution of vertices on the clusters. Such a prior favors a structure with a few populated clusters and several smaller clusters and penalizes balanced clustering models. Our approach overcomes this issue owing to its definition.\\

\emph{(iii)} For an image graph with $K_S$ source and $K_T$ target clusters, every distribution of edges on the tri-clusters -- defined as the cross product of both source, target clusters and time segments -- is equiprobable.

\footnotesize
\[p(E_{\mathcal{IG}}(C_S,C_T,I_n) | K_S, K_T, N) = \dfrac{1}{\binom{|E|+K_SK_TN-1}{K_SK_TN-1}}\]
\normalsize

\emph{(iv)}  For a given cluster of source vertices $c_i=\{v_i, i=1..|c_i|\}$(resp. target vertices $c_j=\{v_j, j=1..|c_j|\}$), every distribution of the out-degree (resp. in-degree) on the vertices of the cluster is equiprobable.
 
\footnotesize
\[ p(d^{out}(v_i) | d^{out}(c_i), \{c_i\}) = \dfrac{1}{\binom{d^{out}(c_i) + |c_i| -1}{|c_i| -1}}\]
\[ p(d^{in}(v_j) | d^{in}(c_j), \{c_j\}) = \dfrac{1}{\binom{d^{in}(c_j) + |c_j| -1}{|c_j| -1}}\] 
\normalsize

\paragraph{Likelihood} Once the image graph parameters are specified owing to the prior definition, the likelihood $P(\mathcal{G} | \mathcal{IG} )$ is defined as the most likely way to observe the graph knowing the image graph parametrization. More formally, the likelihood is made up of the following hypothesis on each parameters of the model :\\

\emph{(i)} On the image graph, every way to draw $|e(c_i,c_j,I_n)|$ edges between the clusters $c_i$ and $c_j$, seen as vertices of the image graph, at the time segment $I_n$ with the $|E|$ edges of the graph is equiprobable and equal to :

\footnotesize
\[P(E|C_S,C_T,I)=\dfrac{ \displaystyle \prod_{c_i \in C_S, c_j \in C_T, I_n\in I} |e(c_i,c_j,I_n)|!}{|E|!} \]
\normalsize

\emph{(ii)} For every cluster of source (resp. target) vertices, every way to
  distribute $d^{out}(c_i)$, the out-degree of the source cluster $c_i$,
  (resp. $d^{in}(c_j)$, the in-degree of the target cluster $c_j$), knowing
  the set of vertices it contains and the degree of each of them, is equiprobable.
  
\footnotesize
\[P(V_S|C_S)=\dfrac{ \displaystyle \prod_{v_i \in V_S}d^{out}(v_i)!}{\displaystyle \prod_{c_i \in C_S}d^{out}(c_i)!} \mbox{ ; }P(V_T|C_T)=\dfrac{\displaystyle \prod_{v_j \in V_T} d^{in}(v_j)!}{\displaystyle \prod_{c_j \in C_T} d^{in}(c_j)!}\]
\normalsize

\emph{(iii)} Every distribution of the rank of the edges timestamps is equiprobable within each time segment.

\footnotesize
\[P(T|I)=\dfrac{1}{\displaystyle \prod_{I_n \in I} |I_n|!} \]
\normalsize

The product of the prior and likelihood terms results in a posterior probability, the negative log of which is used to build a criterion. By optimizing it, vertices which edges in/out-coming tend to be similarly distributed on the clusters are grouped, and the time is discretized into time segments where the edge distribution is stationary. This behavior is illustrated in Section~3.

\begin{MODL}[Cost of an Image Graph]
The Image Graph $\mathcal{IG}$, which is the best synthetic representation of a graph $\mathcal{G}$ according to our modelization, minimizes the following criterion :

\small
\begin{align}
c(\mathcal{IG})& = -\log\left[ P(\mathcal{IG})P(\mathcal{G}|\mathcal{IG}) \right]
\label{eq:MODL}
\end{align}
\end{MODL}
\normalsize


The first term of the criterion corresponds to the negative log of the
prior probability and the second one to the negative log of the
likelihood. In the information theory principles, a negative log of
probability amounts to a Shannon-Fano coding length \cite{Shannon1948}. Thus,
the negative log of the prior probability $-\log(P(\mathcal{IG}))$ is the
description length of the image graph. As for the negative log likelihood
$-\log(P(\mathcal{IG}))$, it is the description length of the graph when
modeled by the image graph. Minimizing the sum of these two terms therefore
has a natural interpretation in terms of a crude MDL (minimum description length) principle \cite{Grunwald2007}.
The criterion $c(\mathcal{IG})$ provides an exact analytical formula for the posterior probability of an image graph $\mathcal{IG}$. That is why the design of sophisticated optimization algorithms is both necessary and meaningful. Such algorithms are described in \cite{Boulle2010}. 

The criterion is minimized using a greedy bottom-up merge heuristic. It starts from the finest image graph, i.e the one with one cluster per vertice and one interval per timestamp. The merges of source and target clusters and the merges between adjacent time intervals are evaluated and performed so that the criterion decreases. This process is reiterated until there is no more improvement, as detailed in Algorithm~\ref{alg:gbum} \\

\begin{center}
\noindent\begin{minipage}{3in}
\footnotesize
\begin{algorithmic}
	\Require $\mathcal{IG}$ (initial solution)
	\Ensure $\mathcal{IG}^*\mbox{ ; } c(\mathcal{IG}^*) \leq c(\mathcal{IG})$
	\State $\mathcal{IG}^* \gets \mathcal{IG}$
	\While{$\mbox{solution is improved}$}
		\State $\mathcal{IG}' \gets \mathcal{IG}^*$
		\ForAll {merge $m$ between 2 source or target clusters or adjacent time segments}
			\State $\{$Consider merge $m$ for Image Graph $\mathcal{IG}\}$
			\State $\mathcal{IG}^+ \gets \mathcal{IG}^*+m$
			\If {$c(\mathcal{IG}^+)<c(\mathcal{IG}')$}
				\State $\mathcal{IG}' \gets \mathcal{IG}^+$
			\EndIf
		\EndFor
		\If {$c(\mathcal{IG}')<c(\mathcal{IG}^*)$}
			\State $\mathcal{IG}^* \gets \mathcal{IG}'$ (improved solution)
		\EndIf
	\EndWhile
\end{algorithmic}
\captionof{algorithm}{\small Greedy Bottom Up Merge Heuristic}\label{alg:gbum}
\end{minipage}
\end{center}

The greedy heuristic may lead to computational issues and a straightforward implementation would be hard to perform. By exploiting both the sparseness of the time-evolving graph and the additivity of the criterion, one can reduce the memory complexity to $O(|E|)$ and the time complexity to $O(|E|\sqrt{|E|}\log|E|)$.

The optimized version of the greedy heuristic is time efficient, but it may fall into a local optimum. This problem is tackled using the variable neighborhood search (VNS) meta-heuristic \cite{Hansen2001}, which mainly benefits from multiple runs of the algorithms with
different random initial solutions. 

\subsection{Simplifying the Image Graph}

When huge graphs are studied, the number of clusters of vertices and of time segments may be too high for an easy interpretation. This problem has been raised in \cite{White1976}, where an agglomerative method is suggested as an exploratory analysis tool. 

The method we propose here consists in merging successively the clusters and
the time segments in the least costly way until the image graph is synthetic
enough for an easy interpretation. From an optimal image graph according to
the criterion detailed in Equation~\ref{eq:MODL}, clusters of source vertices,
of target vertices or time segments are merged sequentially. At each step, the
merged clusters (or time segments) are the ones that induce the smallest increase of
the value of the criterion. 

\begin{diffcrit}
Asymptotically - i.e when the number of edges tends to infinity - the variation $\Delta c$ of the criterion when merging $2$ clusters of (source or target) vertices is equal to the Jensen Shannon divergence between the distribution of the edges on the merged clusters. Similarly, the variation of the criterion when merging two time segments is equal to the Jensen Shannon divergence between the distribution of the edges on the cocluster of source and target clusters for the merged time segments. 

\footnotesize
\begin{align}
\Delta c(\cup (c_1, c_2)) &=(|c_1| + |c_2|) JS^{\alpha_1,\alpha_2}(P_1,P_2) \\
&= (|c_1| + |c_2|) \left( \alpha_1 KL(P_1 || P_{1 \cup 2 }) + \alpha_2 KL(P_2 || P_{1 \cup 2 }) \right) \nonumber
\label{eq:KL}
\end{align}
\normalsize
\end{diffcrit}

where $c_1$ and $c_2$ are the clusters (or time segments) to be merged into a cluster (or time segment) $c_{1 \cup 2}$.  
$P_1$, $P_2$ and $P_{1 \cup 2}$ are the respective distributions of $c_1$, $c_2$ and $c_{1 \cup 2}$. In case of a merge of source clusters:

\footnotesize
\[P_{i \in \{1,2\}}=\left\{\dfrac{|e(c_i,c_j,I_n)|}{|E|} \right\}_{c_j \in C_T, I_n \in I} \]
\[P_{1 \cup 2} = \alpha_1 P_1 + \alpha_2 P_2 \phantom{xx} \alpha_{i \in \{1,2\}} = \dfrac{|c_i|}{|c_1|+|c_2|}
\]
\normalsize

$JS$ is the general Jensen-Shannon Divergence \cite{Lin1991} and $KL$, the Kullback-Leibler Divergence. The full proof is left out for brevity and relies on the Stirling approximation: $\log(n!)= n \log(n) - n + O(\log(n))$, when the difference between the criterion values after and before the merges is computed.

The Jensen-Shannon divergence has some interesting properties: it is a
symetric and non-negative measure between two probability distributions and
the Jensen-Shannon divergence of two identical distributions is equal to
zero. While this divergence is not a metric, as it is not sub-additive, it has
nevertheless the minimal properties need to be used as a similarity measure
within an agglomerative process \cite{Dhillon2003}.

To handle the coarsening of the model, a measure of informativeness of the model is computed at each agglomerative step. It corresponds to the percentage of informativity the model has kept after a merge, compared to a null model.

\begin{informativity}[Informativity of an image graph]
The null model $\mathcal{IG}_{\emptyset}$ is the parametrization of the image graph, such as there is one single cluster of source and target vertices and one time segment. The null model is the synthetic representation of a stationary graph with no underlying structure. Given the best image graph $\mathcal{IG}^*$ obtained by optimizing the criterion defined in Definition~1, the informativity of an image graph $\mathcal{IG}$ is :
\small
\[
\tau(\mathcal{IG})=\dfrac{c(\mathcal{IG})-c(\mathcal{IG}_{\emptyset})}{c(\mathcal{IG}^*)-c(\mathcal{IG}_{\emptyset})}
\]
\normalsize
\end{informativity}
By definition, $\tau(\mathcal{IG}) \leq 1$ ; note that $\tau(\mathcal{IG}) < 0$ is possible when $\mathcal{IG}$ is an irrelevant modelization of the graph $\mathcal{G}$ (e.g. $\mathcal{IG} \neq \mathcal{IG_{\emptyset}}$ when $\mathcal{G}$ is a random graph).
\section{Experiments on Artificial Datasets}
\label{sec:expe}
Experiments have been conducted on artificial datasets in order to investigate the properties of our approach. To that end, we generate artificial graphs with known underlying evolving structures.

\subsection{Experiments on graphs with significative patterns}

\begin{figure}[ht!]%
\centering
 \subfloat[$[0,20[$]{\includegraphics[width=0.11\textwidth]{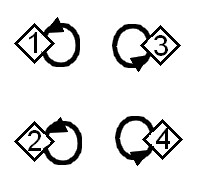}}
 \quad
 \subfloat[$[20,30[$]{\includegraphics[width=0.10\textwidth]{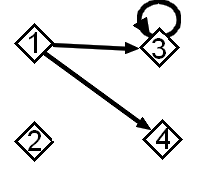}}
 \quad
  \subfloat[$[30,60[$]{\includegraphics[width=0.11\textwidth]{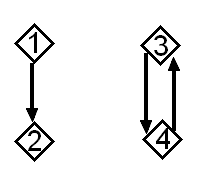}}
  \quad
  \subfloat[$[60,100{]}$]{\includegraphics[width=0.10\textwidth]{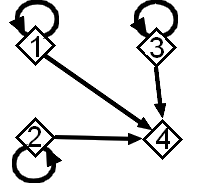}}
 \caption{\small Image graph for each time interval}%
\label{fig:synth}
\end{figure}

The synthetic dataset consists in 40 vertices and a variable number edges. The
vertices are grouped into 4 clusters. There are 5 vertices in clusters $1$ and
$2$, 10 vertices in cluster $3$ and 20 vertices in cluster $4$. The studied
time interval $[0,100]$ has been split into 4 intervals ($I_1=[0,20]$,
$I_2=[20,30]$, $I_3=[30,60]$ and $I_4=[60,100]$) to which are associated
specific image graphs (see Figure \ref{fig:synth}). 
 
The dataset is generated by drawing edges between the vertices according the following process:
\begin{enumerate}
\item A source vertex (and its associated source cluster) and a timestamp are selected uniformly at random.
\item The timestamp is associated to the corresponding time interval which
  gives an image graph specified on Figure~\ref{fig:synth}. 
\item If the source cluster is connected to targets clusters in the image
  graph, then a target vertex is chosen uniformly at random in the union of
  those target clusters and an edge is generated from the source vertex to
  this target vertex.
\end{enumerate}
Then, $30 \%$ of the edges are rewired uniformly at random in order to
  introduce some noise in the dataset. It should be noted that this procedure
  generates numerous multiple edges that can be considered as integer weighted
  edges. 


This synthetic dataset has been generated with a varying number of edges. In
order to obtain reliable results, one hundred graph realisations have been
built for each number of edges. 

\begin{figure}[ht!]%
\centering
  \includegraphics[width=0.35\textwidth]{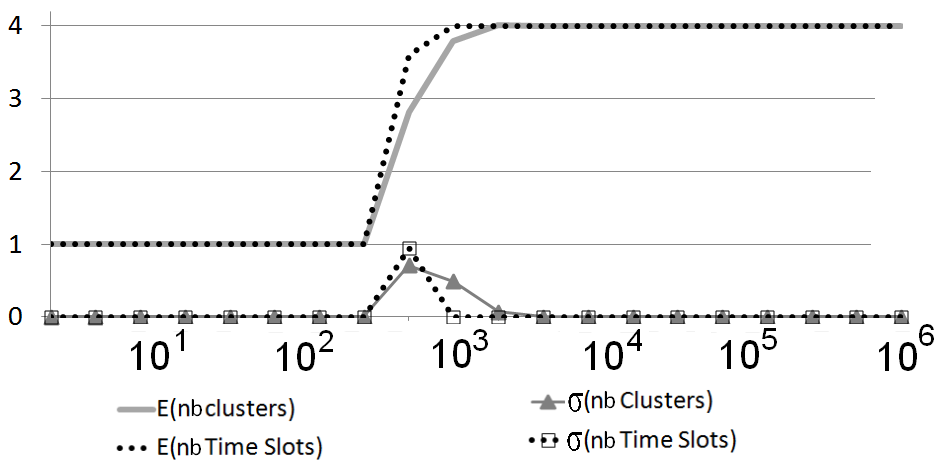}
 \caption{\small Synthetic graph with a time evolving structure. The two top curves are the average numbers of clusters and intervals functions of the number of generated edges. The two bottom curves are the respective standard deviations.}%
\label{fig:pattern}
\end{figure}

The results are displayed on Figure~\ref{fig:pattern}. For a low number of
edges (below 512), the method does not retrieve any cluster. The number of
instances is too low for the method to retrieve any reliable cluster. In this
case, the prior is stronger than the likelihood and the patterns are not
retrieved. Between 512 and 2048, data are numerous enough to detect the
biggest clusters (3 and 4), other ones are sometimes too small and considered
as noise. Finally, beyond 2048 edges, the data are in a large enough amount
and the patterns are retrieved by the method. It is empirically observed that
the numbers of clusters and intervals tend to the true numbers of
patterns. Provided that no more evolution takes place, the method asymptotically estimates the true underlying (evolving) edge distribution.

\subsection{Experiments on stationary graphs}
A stationary graph is a graph whose regular structure (edge distribution) does not evolve over time. To generate such a graph, we have taken the same hundred previous graphs for a given number of edges and we have randomly shuffled the timestamps of the edges. By doing this, we obtain one single new distribution of edges between the clusters that corresponds to a mixture of the previous four distributions. In brief, the graph can be considered as static and the image graph has only one snapshot that is much more complex.

\begin{figure}[ht!]%
	\centering
  \includegraphics[width=0.35\textwidth]{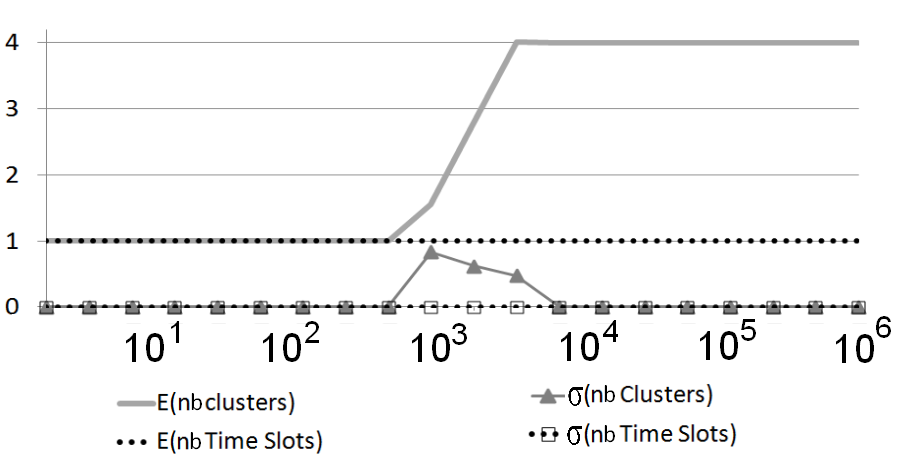}
 \caption{\small Synthetic graph with a stationary structure. The two top curves are the average numbers of clusters and intervals functions of the number of generated edges. The two bottom curves are the respective standard deviations.}
\label{fig:shuffle}
\end{figure}

Figure~\ref{fig:shuffle} shows that whatever the number of edges, the method does not discretize the temporal variable. It is resilient to noise and does not create any spurious time segment. Let us notice that the method requires more data to retrieve the 4 clusters of vertices than the previous experiment. This is due to the complexity of the structure obtained after having mixed the edges distributions of the 4 time segments of the previous experiment.

\subsection{Experiments on Random Graphs}

To obtain random graphs, we have kept the stationary graph and randomly rewired all the edges. By doing this, there is neither underlying structure between clusters of vertices nor temporal structure. Here, the method retrieves one group of vertices and one time segment, with a zero variance on all the hundred graphs whatever the number of edges. The method avoids overfitting and does not detect any spurious structure.

\section{Experiments on a Real-Life Dataset}
Experiments on a real-life dataset have been conducted in order to illustrate the effectiveness of the method on a practical case.
\subsection{The London cycles dataset}
The dataset is a record of all the cycle hires in the Barclays cycle stations of London between May 31st, 2011 and February 4th, 2012. The Data are available on the website of TFL\footnote{Transport for London, \url{http://www.tfl.gov.uk}}. The dataset consists in 488 stations and 4.8 million journeys. It is modelled as a graph with the departure stations as source vertices, the destination stations as target vertices and the journeys as edges. A timestamp on the edges corresponds to the hire time of the day with a minute precision. Actually, there are 1,440 different hire times per day. In this study, we focus on the time of rental as a temporal variable. 
\subsection{The optimal Image Graph}
By applying our method on this dataset, we obtain 296 clusters of source
stations, 281 clusters of target stations and 5 time segments. This has been
computed in 50 minutes using a maximal memory of 4.5 GB. The majority of the
stations being clustered alone within their own cluster, the segmentation is
very fine-grained but this is not the result of an overfitting. In fact, this is due to the huge number of hires, the distributions of edges coming from/to the vertices are characteristic enough to distinguish them from each other. As for the temporal discretization, there are 5 time segments, that is a decent number to make an interpretation.

\subsection{Simplified Image Graph}

Nearly 300 clusters of stations is huge. It is the reason why, we have simplified the image graph in order to enable an interpretation. The image graph has been simplified using the exploratory post-processing described in Section 2.3.  By merging step by step the clusters and the interval, we have chosen to coarsen the image graph until obtaining $70\%$ of informativity (see Section 2.3). This simplification yields 20 clusters of source and target stations while keeping all the 5 time segments.

\begin{figure}[ht!]%
\centering
  \includegraphics[width=0.45\textwidth]{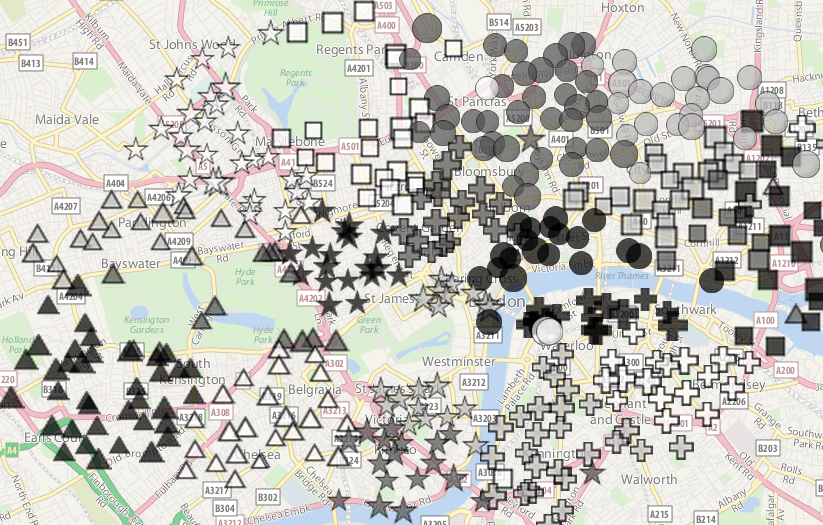}
 \caption{Clusters of source stations plotted on a map of London. There is one greyed symbol per cluster.}%
\label{fig:clust2d}
\end{figure}

A detailed analysis of the clusters reveals that the clustered stations are in general geographically correlated, despite we made no assumption on the proximity of the stations within a cluster. There is also no constraint on our method that yield symetric clusters. In our case, we have considered more interesting to envisage two different clustering structures on target and source stations. Consequently we obtain two different clustering structures on the set of source and target vertices.

Some clusters illustrate this characteristic. First, there is a cluster of stations located within Hyde Park (darkgrey triangles on Figure~\ref{fig:clust2d}). This cluster is exactly the same as source and as target. By contrast, the City of London constitutes one single target cluster while it is split into two source clusters (see Figure~\ref{fig:clust2d}). Like these two examples, most clusters display a strong geographical correlation, except bike stations in front of Waterloo and King's Cross train stations (white circles on Figure~\ref{fig:clust2d}) that have been grouped together while they are very distant. Both being major intercity railroad stations, we can assume that people there have the same behaviour and all converge at the same time to the same point in London, the Central Business District for example.

\subsection{Detailed Visualization}

We use clustering as an exploratory tool and illustrate below the benefit of some specialized visualizations.

\begin{MI2}[Mutual Information between the clusters of stations] This measure quantifies the dependence of two variables. In this first study, the time is left aside and we only focus on the traffic of cycles between the stations all over the day. Let us denote it $MI(C_S,C_T)$, defined as follows~\cite{cover} :

\footnotesize
\begin{eqnarray}
MI(C_S,C_T) &=& \displaystyle\sum_{c_S,c_T} p(c_S,c_T) \log \dfrac{p(c_S,c_T)}{p(c_S)p(c_T)} 
\label{eq:MutualInfo}
\end{eqnarray}
\end{MI2}
\normalsize

Mutual information is necessarily positive. However the involvement to mutual information of a couple of source/target clusters stations can be positive or negative according to whether the observed joint probability of journeys $p(c_S,c_T)$ is above or below the expected probability $p(c_S)p(c_T)$ in case of independence. Displaying such a measure would quantify whether there is a lack or an excess of journeys between two clusters of stations in comparison with the expected number. 

\begin{figure}[ht!]%
\centering
  \includegraphics[width=.45\textwidth]{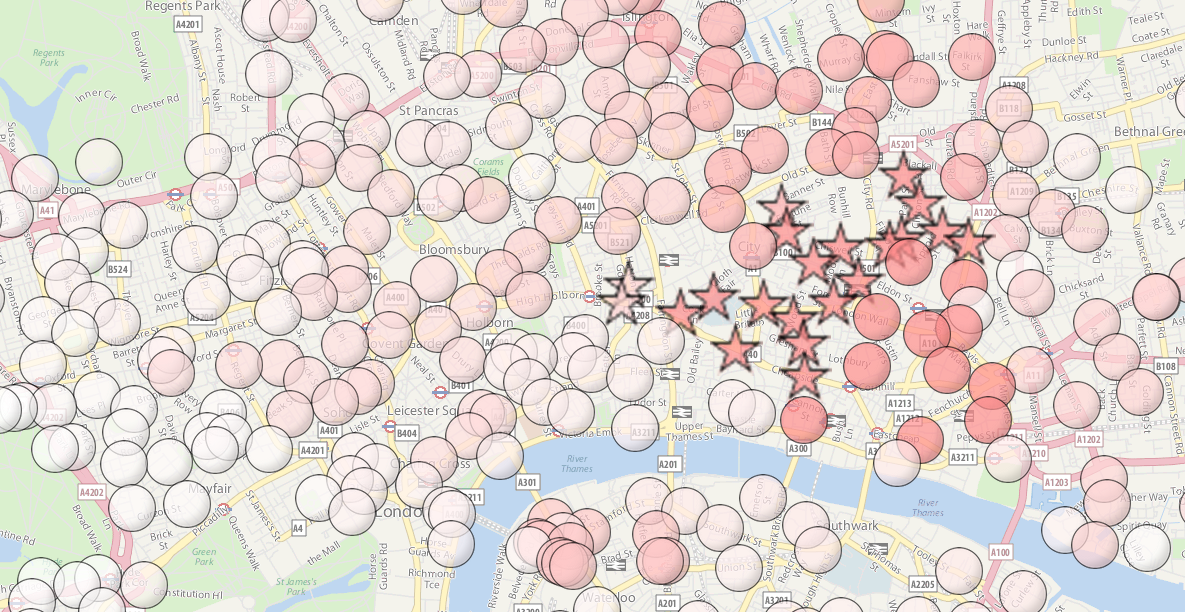}
 \caption{\small Mutual information from the source cluster City/Barbican (stations drawn using stars) to all the clusters. The more a station is colored in red, the more there is an excess of journeys from the stations of the source cluster to the colored station as compared to the expectation under the hypothesis of independance between source and target.}%
\label{fig:mi2d}
\end{figure}

In Figure~\ref{fig:mi2d}, there are clearly more journeys from the cluster "City/Barbican" to the stations of the same neighbourhood, particularly the stations toward the east ; and toward the north, that are mostly residential areas. The conclusions are similar for the majority of the clusters, users tend to make short journeys and leave their bikes in the area where they have hired them.

\begin{MI1}[Mutual Information between journeys and time segments]
We compute the Mutual Information between stations pairs and time segments , $MI[(C_S,C_T),I]$, to study journeys evolution through time. 

\footnotesize
\begin{equation}
MI[(C_S,C_T),I]= \displaystyle\sum_{c_S,c_T,I_n} p(c_S,c_T,I_n) \log \dfrac{p(c_S,c_T,I_n)}{p(c_S,c_T)p(I_n)} \nonumber
\label{eq:MutualInfoCoupleTime}
\end{equation}
\normalsize
\end{MI1}

\begin{figure}[ht!]%
\centering
  \includegraphics[width=.45\textwidth]{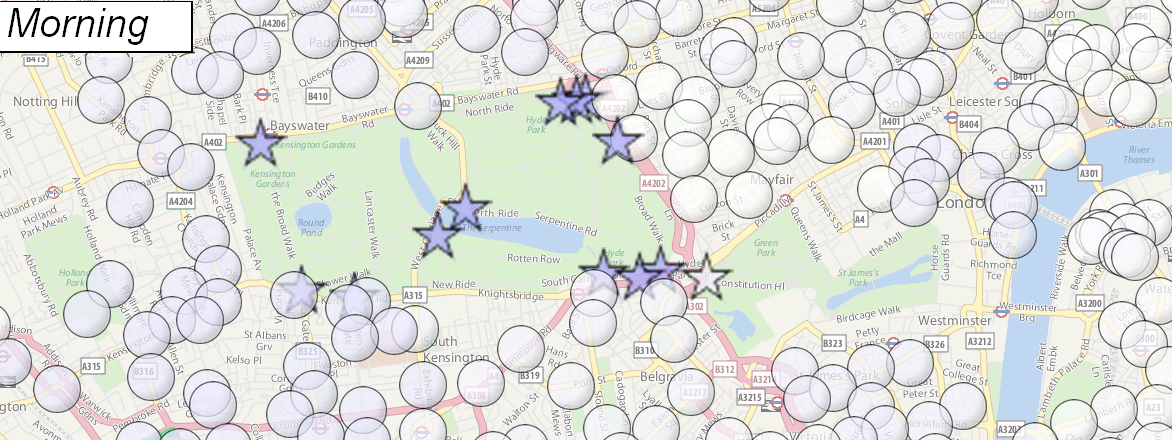} \\[1em]

  \includegraphics[width=.45\textwidth]{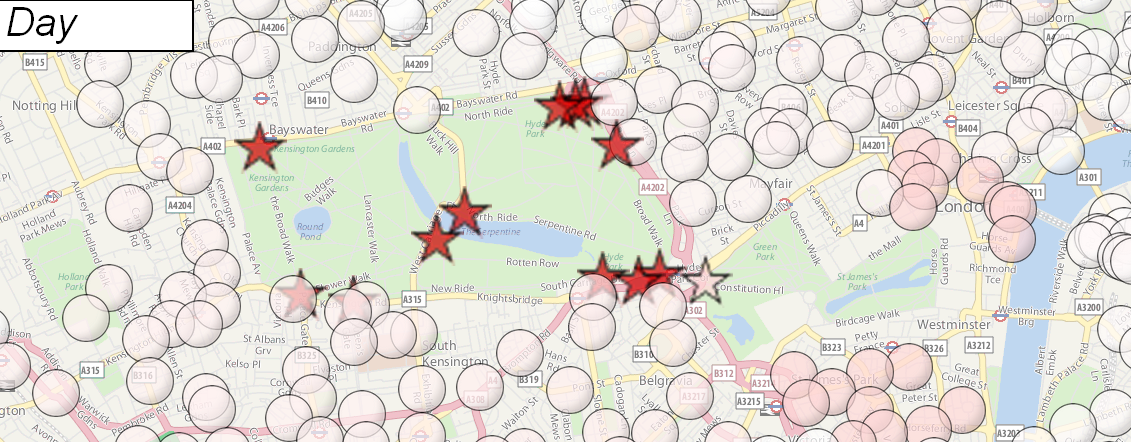}
 \caption{\small Mutual information between the journeys from source cluster Hyde Park (stations drawn using stars) to all the clusters and a time segment, morning (top map) and day (bottom map). The more a station is colored in red (resp. blue), the more there is an excess (resp. lack) of journeys from the stations of the source cluster to the colored station compared to the traffic there is mornings or during the day all over London and traffic there is within Hyde Park all over the day.}%
\label{fig:HPday}
\end{figure}

Similarly to the previous measure, this one aims at showing the couple of clusters between which there is an excess of traffic compared to the usual daily traffic between these stations and the usual traffic at this period in London. To illustrate the measure, we focus on the cluster of Hyde Park and we observe an atypical behaviour. For this cluster, the traffic is lower than what we expected on mornings (see Figure~\ref{fig:HPday}). Indeed, a negative contribution to mutual information means that there is a lack of journeys between the stations of Hyde Park at this time of the day $p(c_S,c_T,I_n)$ compared to the usual traffic between the stations $p(c_S,c_T)$ and the usual traffic at this time segment $p(I_n)$. By contrast there is an excess during the day.\\
\vspace{-0.3cm}

These results are not really surprising because we can assume that in the mornings (time segment from 7:06AM to 9:27AM), when people go to their office, they do not go for a ride in Hyde Park. However, Hyde Park and the morning are both among the busiest clusters and time segments concerning the cycle hires in London. This contrast explains the blue stations on the map of Figure~\ref{fig:HPday}. By contrast, the major part of the traffic within Hyde Park occurs during the day (time segment from 9:27AM to 3:25PM) whereas this time segment corresponds to an off-peak time concerning bikes hires elsewhere in London. This is the reason why Hyde Park is colored in red at the time segment "day" (see Figure~\ref{fig:HPday}). As for the night (time segment from 8:16PM to 4:12AM), all the stations over London are colored in white. Actually, there is the expected number of journeys at this time segment: the Park is closed and the number is so low that the mutual information between journeys belonging from Hyde Park and the time segment "night" is null.

\subsection{Related Works}
A similar study on the bike renting system of Lyon (France) has already been lead \cite{Robardet2009}. However, the dataset needed to be pre-processed to be studied. Indeed the timestamps are aggregated by hour or day depending on which time interval is studied. Moreover, some edges beetween the stations are deleted in order to keep only the "significant" stream of bikes between the stations. Our method does not need such a pre-processing, the time is treated as a continuous variable and the irrelevant streams of bikes between stations are automatically detected as noise by the method.\\
Using the aggregate journey distribution over time to pre-discretize the temporal variable into time segments is conceivable, however making such a choice is not easy and requires an advice of an expert on the studied field, on the one hand to make choices toward the discretization and on the other hand to take a critical look on the results. Because our method simultaneously groups vertices and discretizes the temporal variable without requiring the user to tune parameters, it is particularly suitable for studies on such datasets. All the more so as there is not necessarily a correlation between the journey distribution over time and the evolution of the underlying structures of the graph. Our approach is also able to track very short time segments with significative changes of the structure.


\section{Conclusion}
In this paper, we have dealed with the structural changes within an evolving graph. A novel method, named MODL, aiming at grouping vertices and discretizing the time interval has been introduced. This is related to co-clustering in that we consider the graph as a set of edges described by three features: source vertices, target vertices and time. All of them are simultaneously segmented in order to build a synthetic representation of the graph owing to a set of image graphs that modelizes the static underlying structure of a graph for every time segment. This approach is particularly interesting because it does not require any data preprocessing, such as an aggregation of timestamps or a selection of significative edges. Its good properties have been assessed with experiments on artificial datasets. The method is reliable because it is resilient to noise and asymptotically finds the true underlying distribution. It is also suitable in practical cases as illustrated by the study on the bikes renting system of London. In future works, such a method could be extended to co-clustering in k-dimensions, adding labels on the vertices or another temporal feature (day of week) for example.

\bibliographystyle{IEEEtran}
\small{
\bibliography{biblio}
}

\end{document}